\documentclass{article}
\usepackage[utf8]{inputenc}
\usepackage{xcolor,soul}
\usepackage{cite}
\usepackage{booktabs}
\usepackage{array}
\usepackage{amsfonts, amsmath, amsthm, amssymb}
\usepackage{graphicx}
\usepackage{changepage}
\usepackage{color,soul}
\usepackage{enumitem}   
\usepackage{framed,multirow}
\usepackage{latexsym}
\usepackage{url}
\usepackage{bm}
\usepackage{relsize}
\usepackage{hyperref}
\usepackage{siunitx}
\usepackage{bm}
\usepackage{upgreek}
\usepackage{booktabs}
\usepackage{tabularx}
\usepackage{adjustbox}
\usepackage{amssymb}
\usepackage{pifont}
\newcommand{\xmark}{\ding{61}}%

\usepackage[ruled,vlined]{algorithm2e}

\newcolumntype{L}[1]{>{\raggedright\let\newline\\\arraybackslash\hspace{0pt}}m{#1}}
\newcolumntype{C}[1]{>{\centering\let\newline\\\arraybackslash\hspace{0pt}}m{#1}}
\newcolumntype{R}[1]{>{\raggedleft\let\newline\\\arraybackslash\hspace{0pt}}m{#1}}

\usepackage{amssymb}
\usepackage{pifont}

\usepackage{tikz}

\makeatletter
\newcommand{\thickhline}{%
    \noalign {\ifnum 0=`}\fi \hrule height 1pt
    \futurelet \reserved@a \@xhline
}
\newcolumntype{"}{@{\hskip\tabcolsep\vrule width 1pt\hskip\tabcolsep}}
\makeatother

\usepackage{geometry}
\title{ 
Atlas-Assisted Segment Anything Model for Fetal Brain MRI (FeTal-SAM)
}

\author{Qi Zeng, Weide Liu, Bo Li, Ryne Didier, P. Ellen Grant and Davood Karimi\\
\\
Department of Radiology, Boston Children's Hospital,\\ and Harvard Medical School, USA}

\begin{document}

\maketitle

\begin{abstract}

This paper presents FeTal-SAM, a novel adaptation of the Segment Anything Model (SAM) tailored for fetal brain MRI segmentation. Traditional deep learning methods often require large annotated datasets for a fixed set of labels, making them inflexible when clinical or research needs change. By integrating atlas-based prompts and foundation-model principles, FeTal-SAM addresses two key limitations in fetal brain MRI segmentation: (1) the need to retrain models for varying label definitions, and (2) the lack of insight into whether segmentations are driven by genuine image contrast or by learned spatial priors. We leverage multi-atlas registration to generate spatially aligned label templates that serve as dense prompts, alongside a bounding-box prompt, for SAM's segmentation decoder. This strategy enables binary segmentation on a per-structure basis, which is subsequently fused to reconstruct the full 3D segmentation volumes. Evaluations on two datasets, the dHCP dataset and an in-house dataset demonstrate FeTal-SAM's robust performance across gestational ages. Notably, it achieves Dice scores comparable to state-of-the-art baselines which were trained for each dataset and label definition for well-contrasted structures like cortical plate and cerebellum, while maintaining the flexibility to segment any user-specified anatomy. Although slightly lower accuracy is observed for subtle, low-contrast structures (e.g., hippocampus, amygdala), our results highlight FeTal-SAM's potential to serve as a general-purpose segmentation model without exhaustive retraining. This method thus constitutes a promising step toward clinically adaptable fetal brain MRI analysis tools.\\

\end{abstract}

\section{Introduction}

\subsection{Background and motivation}

The fetal period is arguably the most critical stage in brain development. In-utero magnetic resonance imaging (MRI) has unique capabilities for studying the normal and abnormal development of the brain in this period. It can be used to visualize and quantitatively assess the processes that shape the brain in its earliest developmental phases\cite{gholipour2017normative}. As a complement to ultrasound imaging, fetal MRI has also been established as an indispensable non-invasive tool for diagnosis of congenital disorders and abnormalities \cite{tarui2024fetal}. Constant progress in acquisition protocols and data processing techniques, such as retrospective motion correction and super-resolution methods, have resulted in significant improvements in image quality \cite{rousseau2006registration,snoussi2025haitch,xu2023nesvor}. With the increasing capabilities of fetal brain MRI, it is anticipated that its use in studying brain development in utero will continue to expand. 

Segmentation of specific structures in the fetal brain is essential for quantitative analysis of neurodevelopment because the size and shape of various structures can be strong indicators of normal or abnormal development \cite{ciceri2023review}. Manual segmentation is time-consuming, costly, and may lack reproducibility. Automatic segmentation methods, on the other hand, can compute segmentations with superior speed and reproducibility. With the growing popularity of fetal brain MRI in medical and research applications, automatic segmentation methods are needed to facilitate the processing of large datasets\cite{khalili2019automatic}. As a result, there have been substantial efforts by the research community to develop automatic fetal brain tissue segmentation methods. Recent studies have shown that deep learning techniques excel at this task\cite{zalevskyi2026advances,payette2023fetal}.

However, existing methods suffer from two important limitations. First, the trained model is restricted to a fixed set of region/label definitions. On the other hand, label definitions vary greatly between datasets, and the desired target regions to be segmented depend on the application \cite{dou2020deep}. Hence, with the standard deep learning models, every time the label definitions change, a new training dataset will be needed and a new model will have to be trained. Second, there is a lack of understanding of the reliability of automatic segmentations. Specifically, we do not know to what extent the segmentations computed by these methods are based on the underlying image contrast \cite{karimi2023learning,karimi2022improving}. Moreover, since the reference labels that are used to train and evaluate these models are typically generated with the help of other automatic segmentation methods, there is a lack of critical assessment of the results of deep learning methods and how they may be affected by the biases in the training data \cite{kebiri2024deep,bagheri2025mri}.

This study aims to address these issues. First, we propose a method that (in principle) can segment any structure in the fetal brain. The new method is based on the recent concept of foundation models that go by the name ``segment anything models''. Hence, we refer to the proposed method as Fetal Segment Anything Model (Fetal-SAM). We propose to use atlases/templates to prompt these models to segment the structures of interest.  Second, we analyze the segmentation performance of our proposed method and compare it with the state of the art deep learning models. Our analysis reveals important biases in the results generated with the existing methods. It shows that the segmentations generated with existing methods should be interpreted with care since they may be driven by the biases in the training labels rather than genuine image contrast.

\subsection{Related Works}

\subsubsection{Segmentation of fetal brain tissue in MRI}

Classical non-machine learning segmentation methods, such as multi-atlas techniques, have been applied for fetal brain segmentation \cite{kuklisova2012reconstruction,iglesias2013unified}. However, these methods often suffer from inherent limitations. Multi-atlas methods, for example, are restricted by irreducible registration errors that can lead to incorrect segmentation of intricate structures such as the cortical plate. Studies published in recent years have unequivocally shown that deep learning methods surpass classical segmentation methods for fetal brain segmentation. A number of studies have proposed deep learning methods to segment specific structure, such as the cortical plate \cite{dou2020deep}, or the entire brain tissue\cite{khalili2019automatic}. Special neural network architectures, loss functions, and training procedures have been proposed to improve the segmentation performance \cite{karimi2022improving,karimi2023learning,karimi2025detailed}. Overall, these efforts have led to promising results for automatic fetal brain tissue segmentation in MRI. However, they also present important limitations. Two of these limitation that are the main focus of this study are described below.

An important shortcoming of all existing fetal brain segmentation methods is that they are restricted to a fixed predefined set of regions. This limitation can be crucial because the specific structures of interest vary depending on the application. In some applications, for example, one would like to segment the white matter tissue as a whole, whereas in other settings one may need to segment specific structures in the white matter such as the corpus callosum. Payette et al. use a labeling scheme that divides the entire brain (including the cerebrum, brainstem, and cerebellum) into seven region \cite{payette2021automatic}, while BOUNTI uses 19 regions \cite{uus2023bounti}. Another study has considered 33 regions for fetuses below 32 gestational weeks and 31 regions for fetuses above 32 weeks. More detailed labeling schemes include a larger number of specific structures, such as thalamus, corpus callosum, and lateral ventricles that can be missing in the less detailed schemes\cite{gholipour2017normative}. On the other hand, more specific structures may lack adequate visibility on individual fetal scans. Overall, this limitations means that existing deep learning models are restricted to one specific set of label definitions. If the desired structures change, a new model needs to be trained with a differently-labeled training dataset.

Another persistent challenge for all prior studies is the lack of reliable ground truth segmentation labels for model development and validation. Since manual annotation is prohibitive, it is common to use automatically-generated labels, albeit after verification by human experts, to train and validate the methods \cite{isensee2021nnu}. This approach can lead to significant systematic errors and biases in the method validation, which has not been adequately addressed in prior works\cite{dou2020deep}. Deep learning models have a large field of view that can easily extend to the entire image volume. Consequently, they are capable of learning the spatial arrangement of different brain structures and reproduce this arrangement on the test images. In other words, these models are prone to \emph{duplicating} the patterns observed in the training data even when the local image contrast does not support the segmentation of some of the structures \cite{payette2023fetal}. In order to understand the true merits and capabilities of automatic segmentation methods, it is necessary to know whether the computed segmentations are driven by the underlying image contrast or merely reflect the spatial arrangement of different structures learned from the training data. Such an understanding is essential for proper interpretation of the results generated by automatic segmentation methods.

\subsubsection{Foundation models for image segmentation}

Ever since the success of the original SAM in natural image processing tasks~\cite{kirillov2023segany}, researchers have been exploring the potential of this new paradigm for medical image analysis \cite{SAM4MIS}. Examples of well-known methods include the fine-tuned models presented in~\cite{ma2024segment, cheng2023sammed2d}, where the SAM was extensively tuned with large medical image collections spanning various tissue structures and multiple imaging modalities. These approaches demonstrate promising performance on 2D segmentation tasks, particularly, when user-provided manual prompts are used to aid in tissue structure localization. When fine-tuning SAM for target- and domain-specific tasks, fully automatic prompt generation has been explored through customized prompt encoders designed to localize the tissue of interest~\cite{xie2025self}. Alternatively, methods such as the one in~\cite{ProtoSAM} employ specialized encoders to estimate pseudo-label prototypes, which act as dense mask prompts to guide SAM-based segmentation. Earlier results from these methods suggest that, with proper customization, SAM can be adapted to a fully automatic end-to-end segmentation pipeline.
With the aim of obtaining more accurate label prototypes, atlas-assisted image pre-segmentation was also attempted in~\cite{fan2025maSAM}, where registered atlas label maps were used as dense mask prompts for the SAM decoder. The results show that the adaptation of atlas priors could significantly improve SAM's segmentation performance. However, the atlas label registration setup in this work required a pre-trained CNN to generate a coarse segmentation map, which is then used to initialize the alignment. 
For 3D medical image segmentation, some approaches combine 3D target ROI estimation for multi-slice 2D prompt initialization and learned label fusion for 3D segmentation refinement~\cite{SAM3D}. 
Haoyu \emph{et~al.} have also attempted to train a fully 3D SAM from scratch~\cite{SAM-Med3D}, which requires extensive computational resources and a large dataset. 
Although most existing SAM-based methods show promising performance in their targeted medical segmentation tasks, few have explored SAM’s ability to segment the complex tissue structures encountered in MR neuroimaging. Efforts often concentrate on tumor segmentation and margin assessment, leaving SAM’s performance in detailed brain tissue segmentation underrepresented in the literature. In this work, we bridge this gap by focusing on fetal brain MR image segmentation.

\section{Methods}

Our method builds upon Med-SAM \cite{ma2024segment} and adapts it for segmenting the complex tissue structures in fetal brain MRI. To achieve superior performance, we leverage spatiotemporal fetal brain atlases as priors to guide the segmentation pipeline.
Given a fetal brain image volume, our approach uses age-matched and registered multi-atlas to generate a learned dense prompt, which serves the Med-SAM decoder as additional guidance.
We designed the learned mask prompt to be regressed from two input sources. 
One is the embeddings from the registered multi-atlas images. 
The second is the registered multi-atlas label templates where the target tissue structures are selected and extracted on the image atlas.
In this way, atlas priors in the forms of image features and tissue structure labels can be incorporated into a unified pipeline.
\par
In addition to improved segmentation performance, another key advantage of our proposed pipeline is its flexibility in using prompt templates.
Because the pipeline leverages a registered atlas with tissue label templates for prompt generation, it can be readily adapted to segment any target tissue structure or parcellation definition. 
Moreover, this atlas-based prompting strategy enables multi-class segmentation by simply looping through all targeted structures and repeating the inference with different atlas prompts.
Such flexibility is particularly crucial in fetal brain imaging research, where various parcellation schemes are frequently employed for developmental analyses involving both volumetric assessments. 
Maintaining this versatility is thus highly beneficial for fetal brain imaging studies.
\par
\begin{figure*}[t!]
    \caption{The Proposed Multi-Assisted Driven Fetal Brain SAM Segmentation Framework}
    \label{fig:Network_structure}
    \centering
    \includegraphics[width=1.0\textwidth]{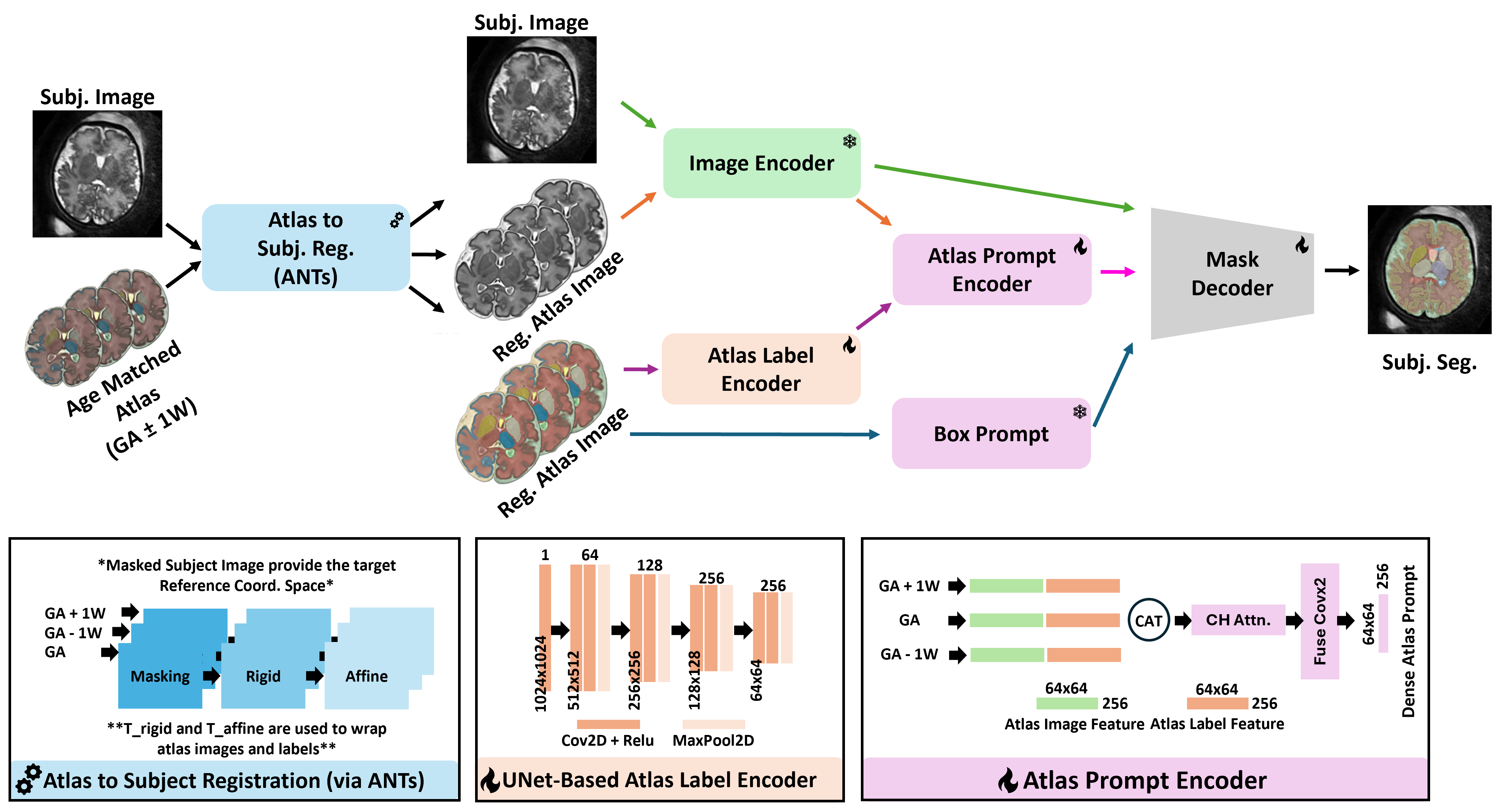}
\end{figure*}

Our proposed multi-atlas driven fetal brain SAM pipeline is graphically summarized in Figure \ref{fig:Network_structure}. 
Architecturally, we follow existing models \cite{ma2024segment} and introduce the necessary modifications to leverage multi-atlas priors.
We start the workflow with multi-atlas to subject image registration. 
With the registered atlas images and the target images, we pass them through the image encoder to extract image feature embeddings. 
We also extract dense features from the registered multi-atlas label template using a trained encoder. 
Then, both the multi-atlas image features and the label features are served to the atlas prompt encoder for dense prompt generation. 
Similar to the Med-SAM configuration, we retain the box prompt, which is computed based on the average label template obtained from the multi-atlas registration results. 
Then, for the segmentation decoder inputs, we have the image feature, the dense-atlas prompt, and the box prompt as shown, respectively. 
Details of all model customizations are presented in the rest of this section.
\par

\subsection{Multi-Atlas to Subject Image Registration}

To ensure that the atlas image with target tissue structure segmentation can be effectively adapted by the Med-SAM model as a source of high-quality priors, accurate spatial normalization between the atlas image space and the subject image space important for network input preprocessing. In this work, atlas-to-subject registration is performed using the Advanced Normalization Tools (ANTS) \cite{ANTS}. In each case, the atlas image is treated as the source image to be registered to the target subject fetal brain MR volume.

As shown at the bottom left of Figure \ref{fig:Network_structure}, registration involves three main processing steps:
\begin{enumerate}
    \item The subject image is first masked to isolate the brain tissue volume.
    \item Rigid registration is then performed for pre-alignment to ensure the atlas image’s brain pose is aligned with the subject image.
    \item Affine registration is applied to optimize the scale and shear.
\end{enumerate}
Through cross-validation on our dataset, we found that using the normalized cross-correlation metric with a local 3D window size of 11$\times$11$\times$11, together with three optimization levels (running \{40, 20, 10\} iterations) and the default multi-level Gaussian smoothing parameters, yielded the most consistent registration results. We therefore retain this configuration for all training, validation, and testing samples in this work.

To ensure the atlas priors can cover a broader range of local variations associated with fetal development at different ages, for each subject image we use $n_{\text{GA}} = 3$ atlas samples whose gestational ages match $\pm 1$ week, thus performing multi-atlas-to-subject registration. For each age, the computed rigid and affine transformations are then used to warp and resample the atlas image and label template to the reference grid provided by the subject image:
\[
I_{\text{prior}} = T_{\text{Affine}}\bigl(T_{\text{rigid}}(I_{\text{Atlas}})\bigr), 
\quad  
Y_{\text{prior}} = T_{\text{Affine}}\bigl(T_{\text{rigid}}(Y_{\text{Atlas}})\bigr).
\]
To serve 2D input images to the Med-SAM encoder, we resample 2D slices of fully aligned image volumes and concatenated as \{$I_{\text{subj.}}, I_{\text{prior-GA}^{-1}}, Y_{\text{prior-GA}^{-1}}, I_{\text{prior-GA}}, Y_{\text{prior-GA}},I_{\text{prior-GA}^{+1}}, Y_{\text{prior-GA}^{+1}} $\},  in axial, coronal, and sagittal directions.
Note that all 2D input slices are resampled and recentered isotopically into 1024$\times$1024 pixels with a spatial resolution of 0.5$\times$0.5 mm$^2$.
\par
\par
\subsection{Atlas Template Encoding}

We propose to modify the standard Med-SAM model to extract features from the registered multi-atlas label template for subsequent dense prompt generation. In this work, the feature extraction encoder we use for the atlas label templates is based on the 2D U-Net architecture, as shown at the bottom of Figure \ref{fig:Network_structure}. We employ \texttt{Conv2D} operations with a 3$\times$3 kernel and \texttt{ReLU} activation to aggregate image features at four different resolution levels $\{512\times512,\,256\times256,\,128\times128,\,64\times64\}$, with channels of $\{64,\,128,\,256,\,256\}$, respectively. Spatial downsampling is performed via max pooling with a 2$\times$2 kernel. This encoder generates learned feature embeddings for all three registered atlas label templates, where each template is encoded as a dense feature map with dimensions $256\times64\times64$. Note that these spatial dimensions are tailored to match the dense prompt input shape of the Med-SAM decoder.
\par
When generating feature embeddings for the registered atlas images, we directly use the pre-trained Med-SAM encoder, keeping its weights frozen throughout training and validation. Since the registered atlas images are reoriented, rescaled, and reshaped samples of average fetal brain MRI data at different gestational ages, we believe that this encoding scheme---without additional encoders---is sufficient. Similar to the atlas label template embeddings, atlas image feature embeddings also have dimensions of $256\times64\times64$ for each of the three gestational age samples.
\par
\subsection{Decoder Prompting with Dense Atlas Prompt and Box Prompt}

After obtaining the feature embeddings of the registered images atlas and the label templates, the second significant modification we propose to the Med-SAM model is the introduction of a dense atlas prompt generator (see the bottom right of Figure \ref{fig:Network_structure}). To ensure that the atlas image features and atlas label features are effectively aggregated and then reshaped, our atlas prompt encoder is based on a channel attention unit and a fusion convolution unit.
\par
First, we spatially concatenate the image feature sets according to the order of the GA groups. Then,
in the channel attention unit, we estimate the channel attention weights via global average pooling, followed by two layers of a channel-wise MLP with \texttt{ReLU} activation. The output channel weights are subsequently normalized using a sigmoid function. After the full feature set is weighted along each channel, it is further spatially aggregated and compressed via the fusion convolution unit, which consists of two \texttt{Conv2D} layers with a 3$\times$3 spatial kernel, an output base channel size of 256, and \texttt{ReLU} activation. Finally, the resulting atlas prompt has dimensions of $256 \times 64 \times 64$, which is used as the dense prompt input to the Med-SAM decoder.
\par

In addition to the dense atlas prompt, we retain the box prompt as in the original Med-SAM configuration. Specifically, we compute the bounding box of the segmentation target by taking the average coordinates of the three registered atlas label maps corresponding to the selected gestational ages. This box prompt plays a crucial role in highlighting the approximate spatial location of the region of interest within the image field of view, thereby guiding the decoder to focus on the relevant anatomical structures. By combining the coarse spatial guidance from the box prompt with the rich prior information embedded in the dense atlas prompt, our customized SAM model preserves both the global layout advantages of the original box prompt approach and the improved segmentation accuracy offered by atlas-driven priors.
\par

\subsection{Handling of over-prompt and under-prompt cases}

\begin{figure*}[t!]
    \caption{Examples of under and over-prompting issues when a registered image atlas was used for target tissue initial localization}
    \label{fig:Prompt_issue}
    \centering
    \includegraphics[width=0.9\textwidth]{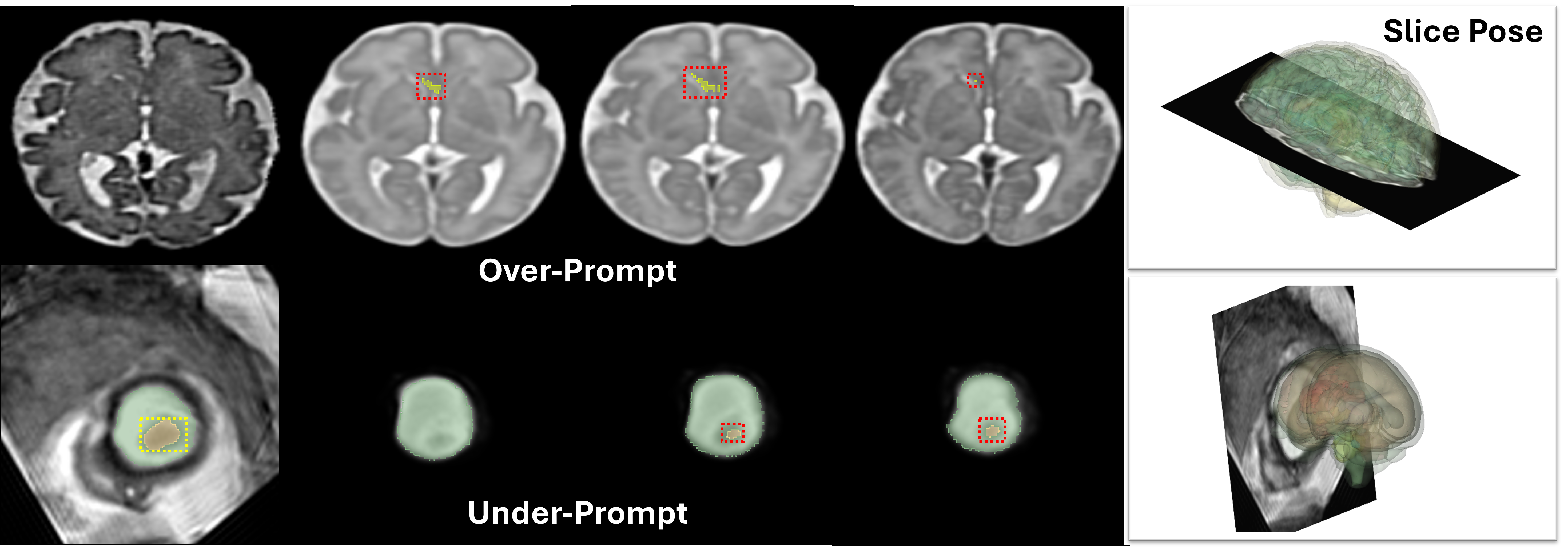}
\end{figure*}

As we are attempting to automate the prompting with atlas guidance, the quality of our prompts relies on accurate atlas to subject image registration. 
Local misalignments between the atlas and subject images can lead to over-prompting or under-prompting, where certain tissue structures are exaggerated or missed. 
This issue is further amplified when volumetric data are resampled into 2D slices, potentially distorting or overlooking smaller anatomical regions, as illustrated in Figure \ref{fig:Prompt_issue}. 
\par
The top row shows a typical example of potential over-prompting in an axial slice, where the basal ganglia appear in all registered atlas label maps (highlighted in yellow) and in the box prompt (highlighted in red).
However, in the subject image shown on the left, there is no clear contrast indicating the presence of the basal ganglia. 
In the original Med-SAM prompt setting, such cases were not considered because the model assumes that all manually selected prompts accurately locate the segmentation target. 
In our pipeline, however, similar situations are relatively frequent due to local registration errors and are included in both training and validation.
Since our model is equipped with both image-based and atlas-based features, we expect that the fine-tuned model can detect and mitigate these over-prompting errors.
\par
Under-prompting, on the other hand, is more difficult to address in our current pipeline, as illustrated by the second example in Figure \ref{fig:Prompt_issue} (bottom row). 
In this case, cortical gray matter (CGM) can be clearly observed in the first sagittal slice of the subject’s left-brain data, where the bounding box is highlighted in yellow. 
From the GA~$-1$~Week atlas in the second column, it is evident that the registration fails to capture the CGM, whereas the matched GA and GA~$+1$~Week results show a clear presence of CGM. 
Consequently, when we prompt for CGM segmentation, the bounding box is computed by averaging only those atlas samples in which the target structure is present. 
In the extreme scenario where all three atlas templates miss the query tissue label class, the prompt can only be refined by improving the registration or resorting to a manual box prompt. 
Fortunately, such cases are quite rare, typically occurring only in the initial slices across the three imaging directions.
\par

\subsection{Implementation Details}

In our implementation, we build upon the Med-SAM-ViT-b model, freezing the original image encoder’s structure and weights while fine-tuning the decoder. Both the atlas label encoder and the atlas prompt encoder are trained from scratch. We adopt a binary segmentation setting in both training and validation, optimizing the network with the loss function:
\[
\mathcal{L} = \mathcal{L}_{\text{Dice}} + \mathcal{L}_{\text{CE}},
\]
where $\mathcal{L}_{\text{Dice}}$ and $\mathcal{L}_{\text{CE}}$ denote the Dice and cross-entropy losses, respectively. All experiments are conducted in PyTorch using an NVIDIA RTX 6000 Ada GPU, and we employ the Adam optimizer with a learning rate of $1\times10^{-4}$. The total number of training epochs varies depending on the dataset in use. 
\par

In the validation and testing phases, we resample each 2D segmentation result back into the original 3D space. 
To capture anisotropic features from multiple perspectives, we repeat the inference on axial, coronal, and sagittal slices. 
Subsequently, we perform label fusion in 3D by combining the segmentation masks from all three orientations using the Simultaneous Truth and Performance Level Estimation (STAPLE) algorithm~\cite{Warfield2004SimultaneousTA}.
This approach consolidates the complementary information obtained from different slice orientations, ultimately producing a more robust and accurate 3D segmentation.
\par

\section{Experiments}

The study has been approved by the Institutional Review Board (IRB) of Boston Children's Hospital. In this subsection, we describe the datasets, evaluation metrics, and baseline methods.

\subsection{dHCP Fetal Brain Dataset:}
The first dataset used in this work was the fetal brain MRI dataset from the The Developing Human Connectome Project (dHCP) \cite{DHCP_dataset}. This dataset includes 297 T2w scans acquired using a 3T Philips Achieva system with a 32-channel cardiac coil, with 273 normal subjects ranging from 21 to 38 weeks GA. Structural T2w imaging involved six different oriented stacks centered on the fetal brain, acquired using a multiband single-shot TSE sequence. The SVR technique presented in \cite{dHCP_SVR} was used to produce isotropic 3D volumes with voxels of size 0.5 mm. A total of 52 cases, equivalent to ~20\% of all subjects, were selected from the data set as ``test subjects'' for the final evaluation. The reference atlas for this dataset was adapted from \cite{DHCP_fetal_atlas}, where atlas construction was completed with the selected cohort that includes 187 fetuses without reported anomalies, from 21 to 37 weeks GA range and with good MRI image quality. Each GA has a high-quality reconstructed mean image atlas with isotropic voxels of size 0.5 mm.
For this dataset, the tissue label considered in this study included the following:
CSF,
Cortical Grey Matter (GM),
Fetal White Matter + Corpus Callosum (WM),
Lateral Ventricle (LV) \xmark,
Cavum septum pellucidum (CSP),
Brainstem (ST),
Cerebellum (CB) \xmark,
Cerebellar Vermis (CBV),
Basal Ganglia (BG) \xmark,
Thalamus (TH) \xmark,
Third Ventricle (V3),
and Fourth Ventricle (V4)
A \xmark next to a label in this list indicates that separate components
in the left and the right brain hemispheres were considered for
that tissue type.

\subsection{CRL Fetal Brain Dataset:}
The second data set used in this work was the collection in house of 294 fetuses with GA between 19.6 and 38.9 weeks (mean 30.6; standard deviation 5.3), which we denote as our CRL fetal brain data set. All images were collected with 3-Tesla Siemens Skyra, Trio, or Prisma scanners using 18 or 30-channel body matrix coils via repeated T2-weighted half-Fourier acquisition single shot fast spin echo (T2wSSFSE) scans in the orthogonal planes of the fetal brain. The slice thickness was 2 mm with no interslice gap, in-plane resolution was between 0.9 mm and 1.1 mm, and the acquisition matrix size was 204×204, 256×256, or 320×320. Volumetric images were reconstructed using an iterative slice-to-volume reconstruction algorithm originally present in \cite{CRL_recon_method}. The resulting 3D images had isotropic voxels of size 0.8 mm. A total of 57 cases, equivalent to 20\% of all subjects, were selected from the dataset as the “test subjects” for final evaluation. The test subjects had GA between 22 and 37 weeks (mean 29.6; standard deviation 3.47). We used the remaining 233 subjects as “training subjects” to develop and train our methods and also to train the competing techniques. For more details on the the dataset preparation, please refer to \cite{More_info_for_CRL_label} and the reference therein. The reference atlas for the CRL-fetal brain dataset was adapted from \cite{CRL_Atlas}. For the GA group from 21 to 38 weeks, each week has a high-quality reconstructed mean image atlas with isotropic voxels of size 0.8 mm, which has been extensively validated.  
For this dataset, the tissue labels considered in this study included the following: 
hippocampus (HP)\xmark, 
amygdala (AM)\xmark, 
caudate nuclei (CD)\xmark,
lentiform nuclei (LN)\xmark, 
thalami (TH)\xmark, 
corpus callosum (CC),
lateral ventricles (LV)\xmark, 
brainstem (ST), 
cerebellum (CR)\xmark,
subthalamic nuclei (SN)\xmark, 
hippocampal commissure (HC),
fornix (FN), 
cortical plate (CP)\xmark, 
subplate (SP)\xmark, 
internal capsule (IC)\xmark, 
and CSF. 

\subsection{Performance Metrics}
In this work, Dice and Jaccard scores are used to evaluate the predicted segmentation, which we compute as:
\[
\bm{\mathrm{DSC}}(p, g) = \frac{2 \sum_{i=1}^{N} p_i \, g_i}{\sum_{i=1}^{N} p_i + \sum_{i=1}^{N} g_i}, \text{and }
\bm{\mathrm{Jaccard}}(p, g) = \frac{\sum_{i=1}^{N} p_i \, g_i}{\sum_{i=1}^{N} p_i + \sum_{i=1}^{N} g_i - \sum_{i=1}^{N} p_i \, g_i}
\]
$g$ and $p$ is used to denote ground truth and the predicted segmentation, respectively.\par
To evaluate the segmentation surface distance error, we first compute the mean surface distance (MSD) error as:
\[
\bm{\mathrm{MSD}}(g, p) = \frac{1}{|S_{g}| + |S_{p}|} \left( \sum_{a \in S_{g}} \min_{b \in S_{p}} \bm{\mathrm{Dist}}(a, b) + \sum_{b \in S_{p}} \min_{a \in S_{g}} \bm{\mathrm{Dist}}(b, a) \right)
\]
where $S$ is the segmentation surface extrapolated from the segmentation label maps. $\bm{\mathrm{Dist}}$ represents operator for computing the 3D Euclidean distance between two surface points. Additionally, the 95 percentile of the Hausdorff Distance (HD95) is computed to assess the large segmentation miss-alignments error:
\begin{align*}
\bm{\mathrm{HD95}}(g, p) = \max\{P_{95\%}(\{& \bm{\mathrm{Dist}}(a, p) : a \in S_{g} \}), \\
&P_{95\%}(\{ \bm{\mathrm{Dist}}(b, g) : b \in S_{p} \})\}.
\end{align*}

\subsection{Baselines}
\begin{itemize}
\item Med-SAM \cite{ma2024segment}: The first baseline method we tested is the Med-SAM model in its ViT-b configuration with pre-trained weights and the box prompt. When preparing the axial, coronal, and sagittal slice inputs for all subjects across the CRL and DHCP datasets, the 3D image volume was first resampled with an isotropic resolution of 1.0 mm$^3$, then 2D images were sliced, centered, and padded into image size of 1024$\times$1024. All box prompts were computed on the basis of ground-truth segmentation. 

\item Med-SAM fine tuned with fetal data \cite{ma2024segment}: The second baseline we tested is the Med-SAM model with a fine-tuned decoder using CRL and DHCP training data. During the fine-tuning, the same 1.0 mm$^2$ 1024$\times$1024 2D image inputs were utilized with pre-computed box prompts for label classes across the two datasets. The decoder weights were tuned with 20 backpropagation echos with images of all training cases.  

\item nnU-Net\cite{nnunet}: When training nnU-Net as our 3D segmentation baseline method, image data from both datasets were resampled with an isotropic resolution of 1.0 mm$^3$, with volume size of 144$\times$144$\times$144. When training the model, both 3D segmentation configurations were trained with a learning rate of 1e-4 with the SGD optimizer and a total of 500 epochs.
For dHCP and CRL datasets, two separate models were trained with the different label class definitions.

\item Swin-UNETR\cite{SwinUnetR}: From recent transformer-based methods, we select Swin-UNETR as the other baseline in our comparison analysis. When training the models, the image input format used was the same as that from nnU-Net. For the dHCP and CRL datasets, two separate models were trained to perform volumetric segmentation in 3D. The Adam optimizer was used with a learning rate of 1e-4 and a total of 500 epochs. When generating the results, the sliding-window inference with label fusion was applied with an overlapping ratio of 75\%. 

\end{itemize}
\section{Results}
\subsection{FeTaL-SAM results}
\begin{figure*}[t!]
    \caption{FeTal-SAM 3D segmentation results comparison to nnUnet and SwinUNetR}
    \label{fig:3D_compare}
    \centering
    \includegraphics[width=1.0\textwidth]{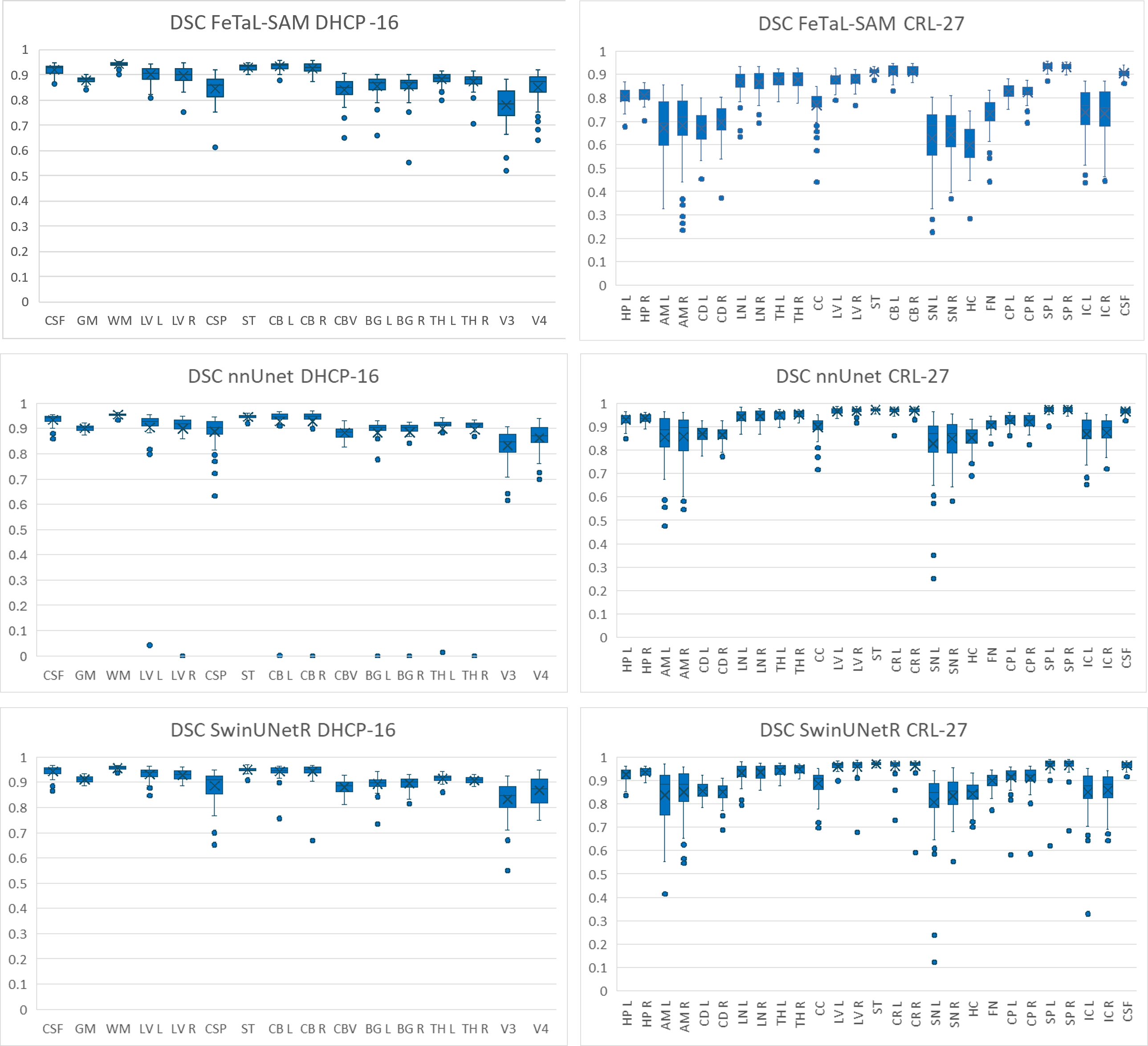}
\end{figure*}
Table \ref{tab:dhcp_crl_results_sam_model} presents our  results with DHCP and CRL datasets using the standard Med-SAM and our proposed FeTal-SAM framework. Results from pre-trained Med-SAM, fine-tuned Med-SAM, and FeTal-SAM are shown in three different rows. As the original Med-SAM was not trained with fetal brain MR data \cite{ma2024segment}, it shows limited segmentation performance for both the DHCP and CRL datasets, where the average DSC scores were reported as 0.252 and 0.393, respectively. When evaluating the segmentation surface distance error in 3D, Med-SAM also reported a high average HD95 above \textgreater 5mm across both datasets, even with box prompts pre-computed based on the ground truth query labels. After Med-SAM was fine-tuned with training cases, improvement of the segmentation performance was observed. The DSC scores for the DHCP and CRL datasets increased to 60\% and 49.3\%, respectively. The overall average HD95 was reduced to \textless 4.0 mm. However, with such a low segmentation accuracy, this fine-tuned Med-SAM is far from acceptable for clinical application deployment. The results from both tests indicate that Med-SAM in its original form has limited capability of segmenting complex tissue structures presented by fetal brain MR data. \\

As Med-SAM model was equipped with our proposed atlas-assisted prompt framework, we see a significantly boosted segmentation performance with the new FeTal-SAM. For the DHCP dataset, the overall average DSC was increased to 88.2\%, and the average HD95 was reduced to 1.04 mm. With the CRL dataset, the average DSC was increased to \textgreater  80\% with the average HD95 at 1.01 mm. Compared to the fine-tuned Med-SAM, nearly 30\% increase of DSC and Jaccard score was achieved, and the average HD95 was reduced to $1/3$. Note that this set of results is generated with a single model. When alternating between the different segmentation target sets, we only need to switch to the age matching image atlas with the paired label template based on different target sets. \\

\begin{table}[]

    \centering
    \caption{FeTal-SAM fine tuning results with DHCP and CRL datasets}
    \label{tab:dhcp_crl_results_sam_model}
    \begin{tabular}{l|cccc|cccc}
    \hline
    & \multicolumn{4}{c|}{\textbf{DHCP}} & \multicolumn{4}{c}{\textbf{CRL}} \\
    \cline{2-9}
    \textbf{Method} & \textbf{DSC} & \textbf{Jaccard} & \textbf{HD95} & \textbf{MSD}
                    & \textbf{DSC} & \textbf{Jaccard} & \textbf{HD95} & \textbf{MSD} \\
    \hline
    Med-SAM       & 0.252 & 0.188 & 9.803 & 5.186 & 0.393 & 0.280 & 5.352 & 1.291 \\
    Med-SAM-FT    & 0.600 & 0.478 & 3.057 & 1.082 & 0.493 & 0.365 & 3.508 & 0.486 \\
    FeTal-SAM     & 0.882 & 0.794 & 1.044 & 0.389 & 0.801 & 0.684 & 1.014 & 0.387 \\
    \hline
    \end{tabular}
\end{table}

\subsection{Comparison to 3D SOTA models}

In Table \ref{tab:dhcp_crl_results_3D}, we compare FeTal-SAM's with 3D SOTA models trained for the DHCP and CRL datasets specifically. For the DHCP dataset, FeTal-SAM's results are really competitive compared to nnUNet and SwinUNetR, with slightly lower DSC which is short of 1.2\% and 2.3\%, respectively. For the Jaccard F1 score, FeTal-SAM is only short of \textless 0.5\%. FeTal-SAM also showed a lower HD95 and MSD compared to nnUNet while slightly fall behind compared to SwinUNetR.
For the CRL data, both nnUNet and SwinUNetR showed a consistent good performance when both of these models are trained and over fitted to this dataset. Across all 27 label classes, both models showed a high overall DSC score above 91\% and a Jaccard F1 score above 84\%. In contrast, our FeTal-SAM model clearly reported a lower DSC and lower Jaccard scores.

To identify the root cause of the lower average DSC, in Figure \ref{fig:3D_compare}, we present test case DSC score distribution across all label classes, where the CRL results are shown in the right column. It is clear to see that FeTal-SAM's results share a similar distributions pattern similar to nnUNet and SwinUNetR. The relatively low average DSC score is mainly caused by the low mean values reported in the HP, AM, CD, SN, HC, and FN classes which are all detailed functional regions around Basal Ganglia and Thalamus. Such results indicate that accurate segmentations of these detailed tissue structures may require detailed 3D spatial prior and features which are easy to learn by the 3D nnUNet and SwinUnetR models which are over-fitted to the CRL and dHCP dataset. Even though FeTal-SAM has equipped with rich priors from the registered age matching image atlas, without the support of 3D spatial prior, carrying out accuracy segmentation details for these structures remains a challenging task for SAM based 2D segmentation methods, where the network will only focus on local image contrast rather than the spatial arrangement of all structures in the whole brain 3D volume.

\begin{table}[b!]
    \centering
    \caption{FeTal-SAM comparison to 3D SOTA models}
    \label{tab:dhcp_crl_results_3D}
    \begin{tabular}{l|cccc|cccc}
    \hline
    & \multicolumn{4}{c|}{\textbf{DHCP}} & \multicolumn{4}{c}{\textbf{CRL}} \\
    \cline{2-9}
    \textbf{Method} & \textbf{DSC} & \textbf{Jaccard} & \textbf{HD95} & \textbf{MSD}
                    & \textbf{DSC} & \textbf{Jaccard} & \textbf{HD95} & \textbf{MSD} \\
    \hline
    nnUNet        & 0.902 & 0.830 & 1.155 & 0.430 & 0.918 & 0.854 & 0.827 & 0.180 \\
    SwinUNetR     & 0.911 & 0.840 & 0.958 & 0.298 & 0.911 & 0.840 & 0.958 & 0.298 \\
    FeTal-SAM     & 0.882 & 0.794 & 1.044 & 0.389 & 0.801 & 0.684 & 1.014 & 0.387 \\
    \hline
    \end{tabular}
\end{table}

\section{Discussion}

\begin{figure*}[t!]
    \caption{Graphical segmentation results from DHCP test cases: A sample case with GA at 22 weeks is shown on the top three rows, and a sample case with GA at 30 weeks is shown at the bottom, respectively.}
    \label{fig:dhcp_example}
    \centering
    \includegraphics[width=1.0\textwidth]{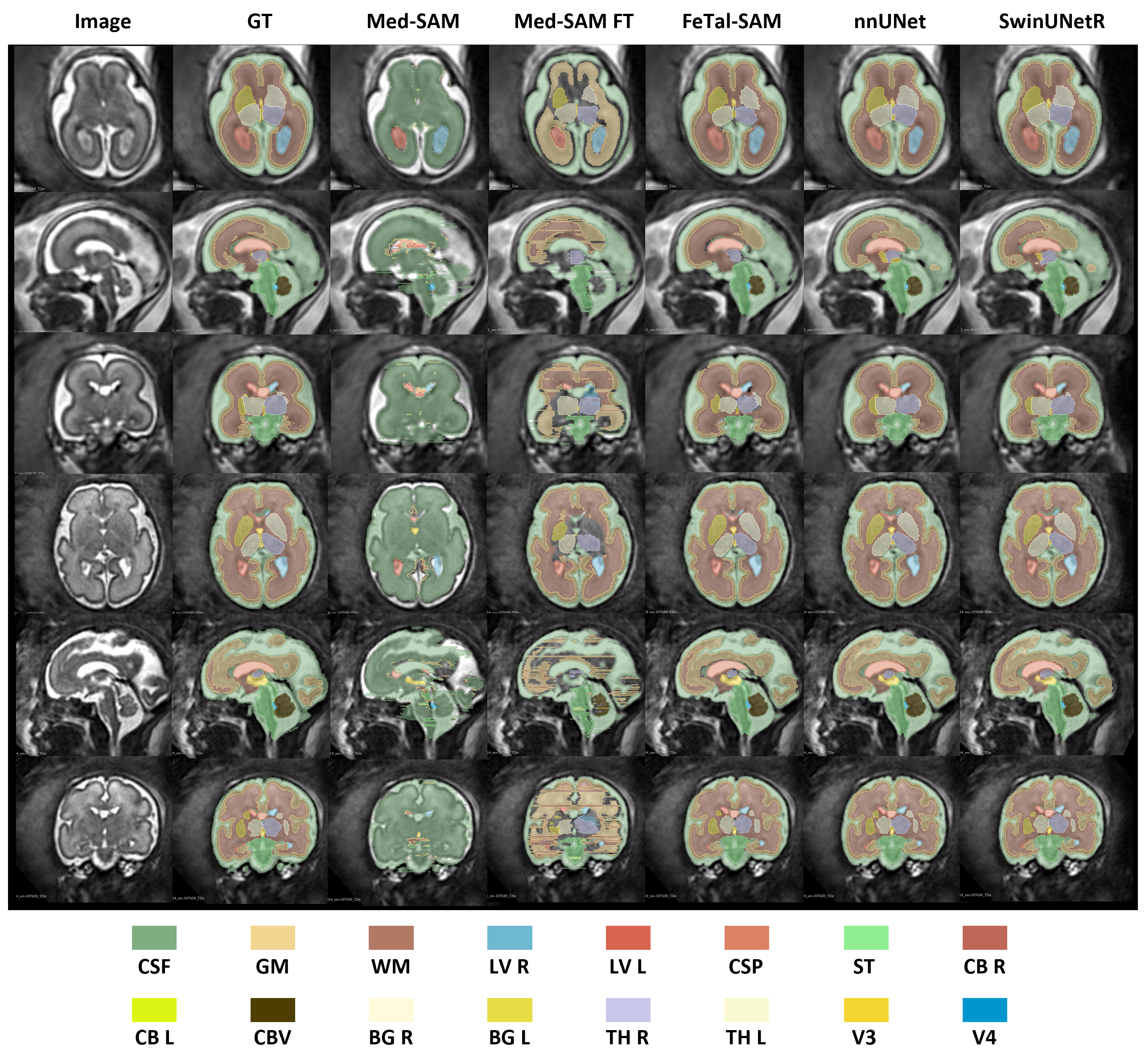}
\end{figure*}

From the validation results across two fetal brain MR datasets and two different label definitions, FeTal-SAM demonstrates a promising performance improvement over the existing Med-SAM setup. By leveraging the registered atlas and target tissue label templates, FeTal-SAM effectively segments complex tissue structures in fetal brain MR data. For tissue structures that present good contrast in the MR images, FeTal-SAM achieves performance comparable to 3D segmentation models specifically trained for each dataset. Figure~\ref{fig:dhcp_example} illustrates segmentation results on the DHCP dataset, comparing the reconstructed 3D segmentation maps for two test subjects at gestational ages 22,W and 30,W. In both cases, FeTal-SAM showed high consistency with the ground truth and closely matched the results from nnUNet and SwinUNetR. Notably, FeTal-SAM still operates in a binary 2D segmentation fashion, inherited from Med-SAM, where target tissue structures are segmented slice by slice based on the registered image atlas and label templates. The STAPLE-based multi-slice label fusion further refines these segmentation results. Compared with other methods, FeTal-SAM exhibits significantly less overestimation.
\begin{figure*}[t!]
    \caption{Graphical Comparison of segmentation errors for different tissue structures with a test sample from our CRL dataset. The left panel shows the results from FeTal-SAM and nnUNet for cortical plate, sub-plate and cerebellum. The right panel shows results of more challenging small tissue structures, including hippocampus and amygdala. The ground truth (GT), over segmentation and under segmentation are displayed in red, green and blue, respectively. }
    \label{fig:crl_nnunet_detailed}
    \centering
    \includegraphics[width=1.0\textwidth]{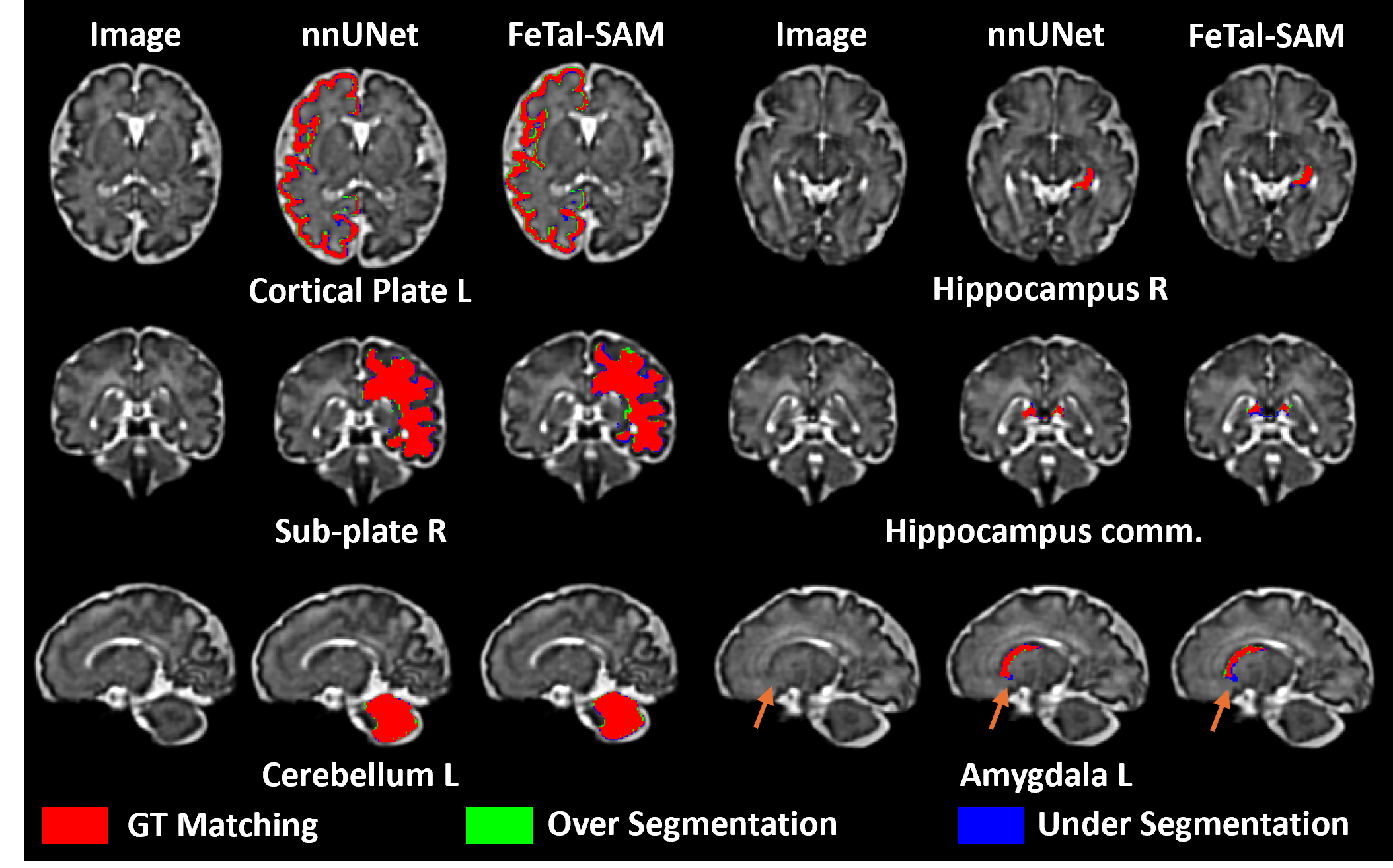}
\end{figure*}
From the earlier analysis in Figure~\ref{fig:3D_compare}, FeTal-SAM lags behind dataset-specific 3D models in segmenting fine details of the basal ganglia and thalamus. Without leveraging priors learned from 3D data, capturing these intricate structures from 2D slices alone can be challenging. Figure~\ref{fig:crl_nnunet_detailed} provides a closer look at this issue by comparing tissue structures labeled by FeTal-SAM and nnUNet in a test case from the CRL dataset. For larger structures with good contrast (e.g., the subplate, cortical plate, and cerebellum), FeTal-SAM produces results comparable to or even better than those from nnUNet. However, for smaller structures located near the basal ganglia and thalamus (e.g., the hippocampus and amygdala), FeTal-SAM exhibits more under- and over-segmentation errors. A detailed inspection suggests that the limited tissue contrast in these regions poses a significant challenge. While 3D learning methods can rely on spatial priors to “inpaint” small structures, FeTal-SAM’s reliance on local tissue contrast in each 2D slice makes it more difficult to accurately capture these hard-to-visualize areas.

\begin{figure*}[h]
    \caption{Graphical Comparison of Image Contrast Profiles and Ground Truth (GT) Labels.Axial slices from two subjects at gestational ages (GA) of 26 weeks (top row) and 32 weeks (bottom row) illustrate the comparison. The left column shows the image slice; the second column presents the label GT; the third column highlights a zoomed-in region with the MR image as the background and the tissue label as an overlay. In the last column, the normalized image intensity profile and the label profile along the red line are plotted. In both examples, the label distribution does not consistently follow the local contrast pattern, and the tissue contrast is relatively low.}
    \label{fig:crl_contrast_issue}
    \centering
    \includegraphics[width=1.0\textwidth]{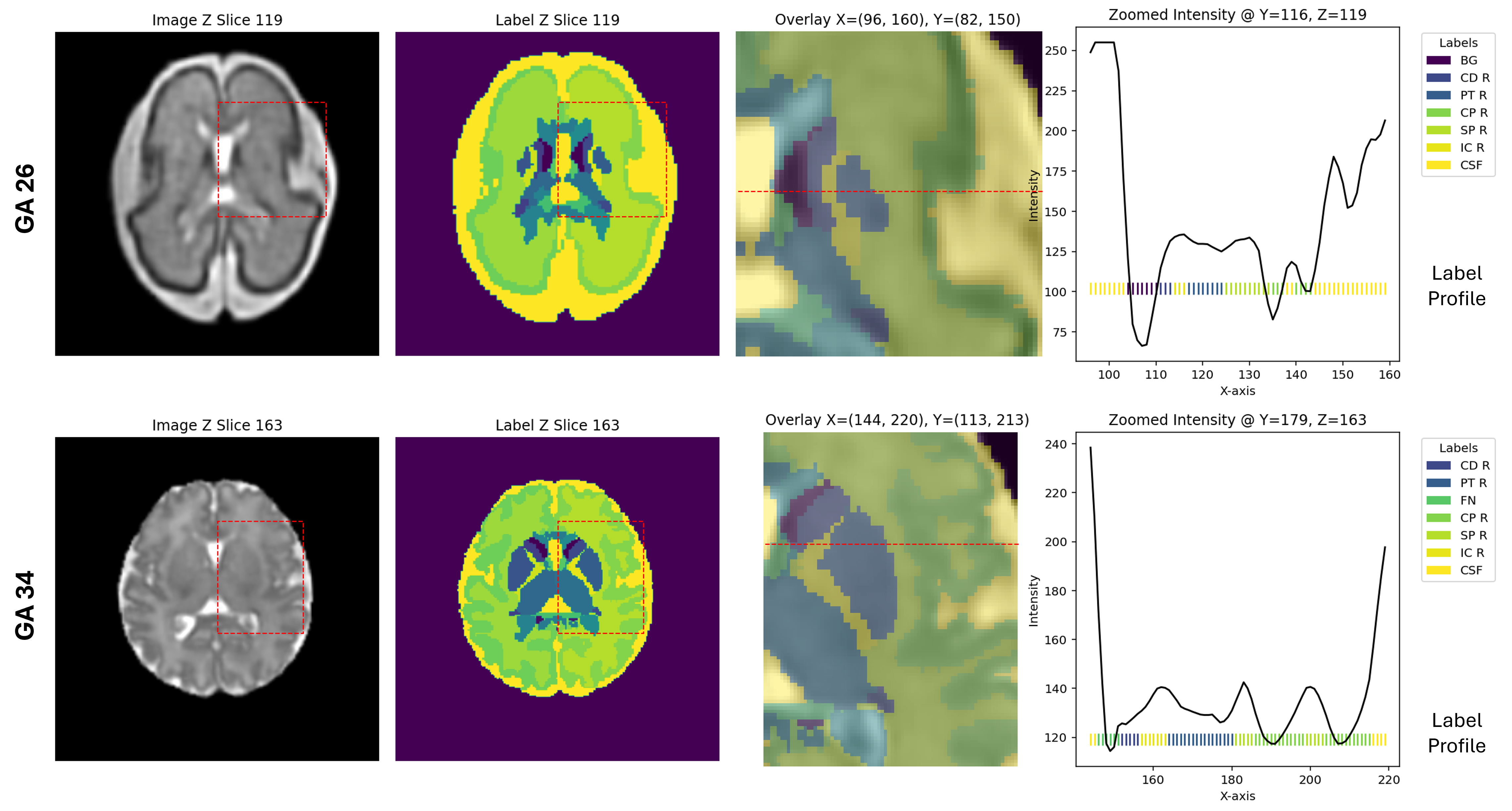}
\end{figure*}

To further highlight the low-contrast challenge surrounding the ganglia and thalamus, we present two test case examples in Figure~\ref{fig:crl_contrast_issue}, showing the central axial slice of two subjects from the CRL dataset at gestational ages 26 and 34. The ground-truth labels used for validation are overlaid, and a single axial line (marked in red) indicates where the relative tissue contrast profile (shown in the right column) was extracted. Various tissue labels appear in different colors were shown as an overlay. The normalized image intensity across the caudate, putamen, and internal capsule reveals relatively limited contrast, making it difficult to confirm whether the labels truly follow the contrast variations. This observation suggests that the detailed ganglia and thalamus labels in this dataset may primarily rely on spatial priors from an average brain atlas. For SAM-based models that derive labels by leveraging image-contrast features, enforcing stronger consistency with atlas-based label priors can be challenging.

To verify the difficulty of labeling these smaller tissue structures, we asked four clinical experts to annotate a subset of four test cases from the CRL data at gestational ages 23, 27, 31, and 35. In each case, five different 2D slices were provided for experts to segment the amygdala, caudate, subthalamic nuclei, hippocampal commissure, and fornix. Table~\ref{tab:dsc_icc2} presents the evaluation results, showing the average Dice Similarity Coefficient (DSC) between each expert’s labels and the current ground truth for each tissue class. The single-random-rater intraclass correlation coefficient (ICC2) was also computed to assess inter-expert variability. Clearly, all four experts struggled to produce labels consistent with the ground truth, with the average DSC for all label classes falling below 60\%. The relatively low ICC2 values further indicate substantial inter-observer variability. One exception is the fornix, which showed a higher ICC2; however, experts still exhibited consistently poor agreement with the ground truth for this structure.

These observations underscore the difficulty of generating reliable segmentations for the basal ganglia and thalamus when tissue contrast is limited in fetal MR data. Although dataset-specific 3D models, which leverage strong spatial priors, may achieve high agreement with training labels, the extent to which these labels align with actual tissue contrast remains uncertain. Such models might be biased toward the learned priors. In contrast, FeTal-SAM provides a potentially more objective approach by using a single model to segment all data based on customizable label definitions.


\begin{table}[ht]
\centering
\caption{Segmentation results from four experts for eight tissue regions}
\label{tab:dsc_icc2}
\scalebox{0.8}{
\begin{tabular}{lcccccccc}
\hline
                & AM L & AM R & CD L & CD R & SN L & SN R & HC   & FN   \\ 
\hline
\textbf{Expert 1 DSC \%($\sigma$)} & 0.46(0.33) & 0.40(0.27)& 0.40(0.28) & 0.38(0.30) & 0.60(0.15) & 0.53(0.09) & 0.36(0.39) & 0.27(0.38) \\
\textbf{Expert 2 DSC \%($\sigma$)} & 0.51(0.23) & 0.59(0.24) & 0.14(0.16) & 0.10(0.21) & 0.27(0.22) & 0.44(0.26) & 0.08(0.16) & 0.23(0.24) \\
\textbf{Expert 3 DSC \%($\sigma$)} & 0.42(0.14) & 0.31(0.22) & 0.31(0.34) & 0.29(0.38) & 0.44(0.15) & 0.34(0.29) & 0.45(0.17) & 0.23(0.32) \\
\textbf{Expert 4 DSC \%($\sigma$)} & 0.26(0.23) & 0.44(0.37) & 0.24(0.38) & 0.25(0.36) & 0.19(0.35) & 0.28(0.36) & 0.46(0.14) & 0.19(0.28) \\
\hline
\textbf{ICC(2)}      & 0.26 & 0.21 & 0.45 & 0.50 & 0.19 & 0.04 & 0.44 & 0.91 \\
\hline
\end{tabular}
}
\end{table}

\section{Conclusion}
In this work, we proposed FeTal-SAM, an atlas-assisted framework that extends the Segment Anything Model (SAM) for fetal brain MRI segmentation. The key innovation lies in leveraging registered multi-atlas label templates as dense prompts, which are fed into SAM’s segmentation decoder alongside bounding-box prompts. By operating on 2D slices and merging results via 3D label fusion, FeTal-SAM enables flexible, on-demand segmentation of any anatomical structure without requiring task-specific retraining or large, exhaustively annotated datasets.
Our experiments on two fetal brain MRI datasets, dHCP and CRL, demonstrate that FeTal-SAM performs comparably to specialized 3D deep-learning models for well-contrasted structures (e.g., cortical plate, cerebellum). At the same time, it maintains superior adaptability, allowing users to alter parcellation schemes simply by providing different atlas-based label prompts. While smaller, low-contrast structures such as the hippocampus or amygdala remain more challenging to segment reliably, our analysis highlights that these limitations often stem from the inherent ambiguities in fetal MRI contrast rather than the model architecture itself.
By unifying atlas-based priors with SAM’s general-purpose segmentation capabilities, FeTal-SAM fosters both flexibility and interpretability in fetal neuroimaging. Future directions include enhancing 3D spatial understanding, refining atlas registration techniques, and incorporating more robust uncertainty quantification. Ultimately, FeTal-SAM sets the stage for versatile, efficient segmentation pipelines that can accommodate rapidly evolving clinical and research demands.\\

\section{Acknowledgements}

This research was supported in part by the National Institute of Neurological Disorders and Stroke and Eunice Kennedy Shriver National Institute of Child Health and Human Development of the National Institutes of Health (NIH) under award numbers R01HD110772 and R01NS128281. The content of this publication is solely the responsibility of the author and does not necessarily represent the official views of the NIH.

\bibliographystyle{unsrt}
\bibliography{references}

\end{document}